\documentclass[sigconf]{acmart}

\usepackage{multirow}
\usepackage{bbm}

\makeatletter
\def\@projectpage{}
\newcommand{\projectpage}[1]{\gdef\@projectpage{#1}}
\usepackage{etoolbox}
\pretocmd{\@printtopmatter}{%
  \ifx\@projectpage\@empty\else
    \global\setbox\mktitle@bx=\vbox{%
      \unvbox\mktitle@bx
      \vskip 6\p@
      \centerline{\Large \url{\@projectpage}}%
      \vskip 12\p@
    }%
  \fi
}{}{}
\makeatother

\AtBeginDocument{%
  }

\renewcommand\footnotetextcopyrightpermission[1]{}

\setcopyright{none}
\acmISBN{}
\acmDOI{}

\begin{document}

\title{AniCrafter: Customizing Realistic Human-Centric Animation via Avatar-Background Conditioning in Video Diffusion Models}

\author{Muyao Niu}
\affiliation{%
  \institution{The University of Tokyo}
  \city{Tokyo}
  \country{Japan}
}
\email{muyao.niu@gmail.com}

\author{Mingdeng Cao}
\affiliation{%
  \institution{The University of Tokyo}
  \city{Tokyo}
  \country{Japan}
}
\email{mingdengcao@gmail.com}

\author{Yifan Zhan}
\affiliation{%
  \institution{The University of Tokyo}
  \city{Tokyo}
  \country{Japan}
}
\email{zhan-yifan@g.ecc.u-tokyo.ac.jp}

\author{Qingtian Zhu}
\affiliation{%
  \institution{The University of Tokyo}
  \city{Tokyo}
  \country{Japan}
}
\email{qtzhu@g.ecc.u-tokyo.ac.jp}

\author{Weihang Ran}
\affiliation{%
  \institution{The University of Tokyo}
  \city{Tokyo}
  \country{Japan}
}
\email{ran-weihang@g.ecc.u-tokyo.ac.jp}

\author{Yanhong Zeng}
\affiliation{%
  \institution{Ant Group}
  \city{Hangzhou}
  \country{China}
}
\email{zengyh1900@gmail.com}

\author{Xiao Sun}
\affiliation{%
  \institution{Shanghai Artificial Intelligence Laboratory}
  \city{Shanghai}
  \country{China}
}
\email{sunxiao@pjlab.org.cn}

\author{Zhihang Zhong}
\correspondingauthor
\affiliation{%
  \department{School of Artificial Intelligence (SAI)}
  \institution{Shanghai Jiao Tong University}
  \city{Shanghai}
  \country{China}
}
\email{zhongzhihang@sjtu.edu.cn}

\author{Yinqiang Zheng}
\affiliation{%
  \institution{The University of Tokyo}
  \city{Tokyo}
  \country{Japan}
}
\email{yqzheng@ai.u-tokyo.ac.jp}

\projectpage{https://myniuuu.github.io/AniCrafter/}

\renewcommand{\shortauthors}{Muyao Niu et al.}

\begin{abstract}
  Recent advances in video diffusion models have substantially enhanced character animation techniques. However, existing methods primarily depend on structural conditions, such as DWPose or SMPL-X, to animate character images, which limits their effectiveness in open-domain scenarios involving dynamic backgrounds or complex character-scene interactions. This study presents AniCrafter, a diffusion-based human-centric animation model designed to seamlessly integrate and animate a given character within open-domain dynamic backgrounds while adhering to specified human motion sequences. Built upon advanced Image-to-Video (I2V) diffusion architectures, the model introduces an innovative ``avatar-background'' conditioning mechanism that reformulates open-domain human-centric animation as a restoration problem, thereby achieving versatile, occlusion-aware animation results. Experimental evaluations demonstrate that the proposed approach outperforms current state-of-the-art methods and exhibits an exceptional capability in handling challenging scenarios. Codes and model are available at: \url{https://github.com/MyNiuuu/AniCrafter}
\end{abstract}

\maketitle

\section{Introduction}
\label{sec:intro}

Recent breakthroughs in generative modeling~\cite{ho2020denoising,rombach2022high,peebles2023scalable}, particularly in Video Diffusion Models (VDMs)~\cite{yang2024cogvideox,blattmann2023stable,xing2024dynamicrafter,he2022latent,chen2024videocrafter2}, have markedly advanced the field of dynamic video generation. Leveraging the strong generative capabilities of these models, numerous studies~\cite{hu2024animate,niu2024mofa,zhang2024mimicmotion,zhu2024champ,yu2024viewcrafter,shi2024motionstone,he2024cameractrl} have explored controllable video generation guided by the prior knowledge embedded within VDMs.

\begin{figure*}
\setlength{\abovecaptionskip}{2mm}
  \includegraphics[width=\textwidth]{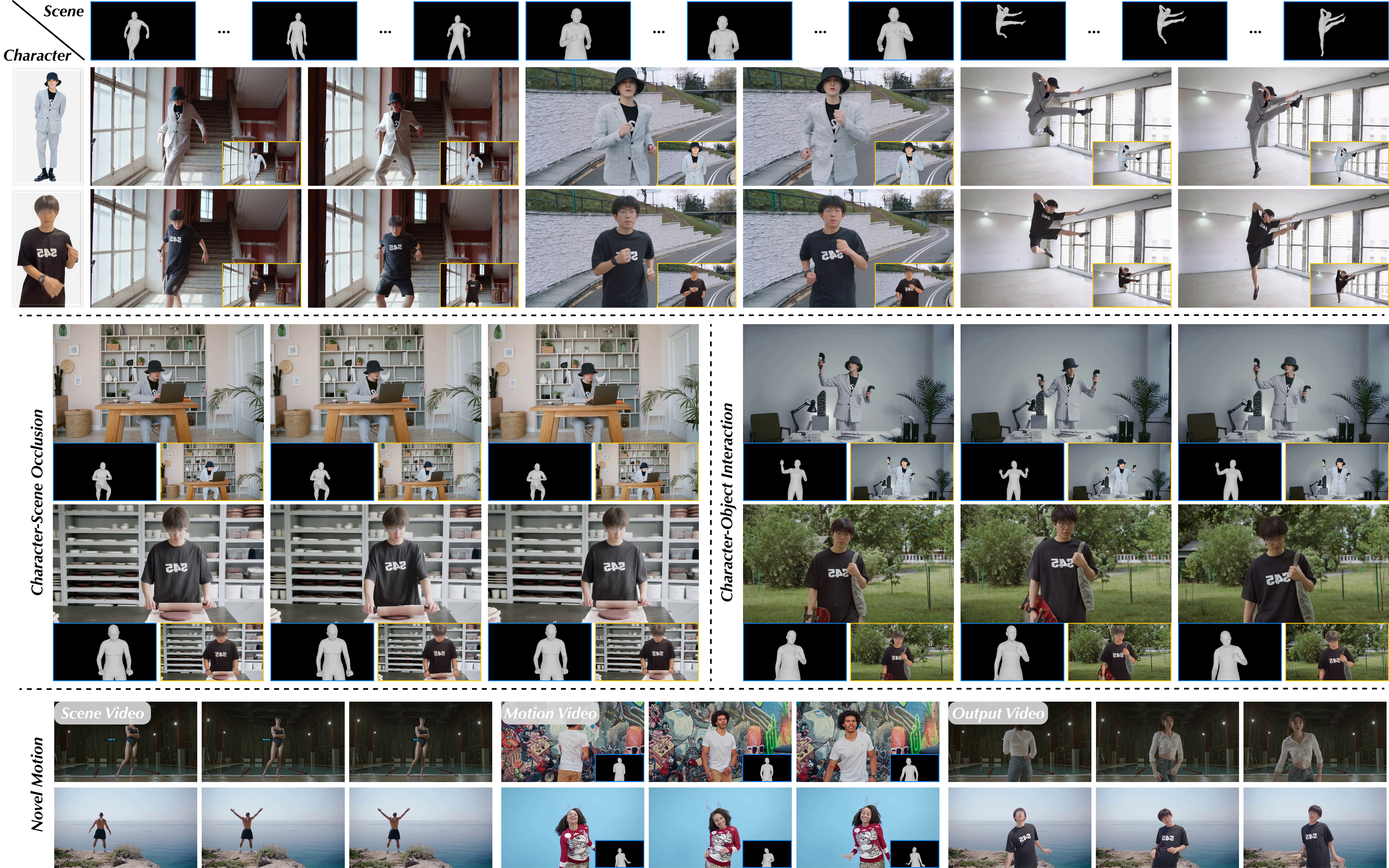}
  \caption{
    Within any open-domain dynamic scene, \textbf{AniCrafter} integrates and animates given characters with harmonized illumination and appearance, following specified motion sequences. In addition to SMPL-X (blue boxes), an ``avatar-background'' conditioning signal (yellow boxes) is introduced to enable robust and diverse human-centric animation. In addition, the model supports \textbf{character-scene occlusion},  \textbf{character-object interaction}, and \textbf{novel motion}. 
  }
  \label{fig:teaser}
\end{figure*}

In the most closely related research domain, a range of approaches \cite{hu2025animate,wang2024unianimate,wang2024disco,xu2024magicanimate,chang2025x,cao2024consistent,lin2025omnihuman,niu2026motion,zhan2025towards,xu2025sequential,xu2025sequential,chen2024within,gong2025visionary,zhang2025diffbody} has been developed to animate static character images into motion sequences defined by representations such as DWPose~\cite{yang2023effective}, DensePose~\cite{guler2018densepose}, SMPL-X~\cite{SMPLX2019}, or 3D skeletons~\cite{men2024mimo}. Beyond animating human body parts, these methods effectively capture natural motion dynamics in non-rigid regions, such as hair or garments, which are difficult to model explicitly with conventional 3D techniques. Despite these strengths, existing approaches typically rely on basic structural conditioning and often exhibit structural distortions or temporal inconsistencies. Furthermore, most methods assume that the input character image and the generated video share the same static background, focusing solely on animating the character itself.

Another technical paradigm~\cite{qiu2025LHM,qiu2024anigs,wen2025lifegom,motionshop2} employs explicit 3D modeling techniques, particularly 3D Gaussian Splatting~\cite{kerbl20233d}, to reconstruct an animatable 3D human avatar from a single reference image. In contrast to diffusion-based approaches that directly synthesize 2D videos, these methods adopt a two-stage pipeline: (1) constructing an explicit 3D human avatar using the segmented character image without the background, and (2) animating and rendering the 3D avatar into 2D image space according to 3D pose transformations such as SMPL-X. This pipeline produces multi-view-consistent outputs with strong structural integrity, owing to the geometric constraints of 3D warping and transformation. However, avatars reconstructed through these methods typically lack high-frequency appearance details and fail to reproduce realistic motion dynamics due to the inherent limitations of explicit 3D modeling, particularly for non-rigid components such as hair or garments. Moreover, the rendered characters often exhibit poor visual harmony when composited directly into background videos.

Building upon the analysis of the two aforementioned paradigms, this study introduces \textbf{AniCrafter}, a human-centric video animation model capable of integrating and animating arbitrary characters within diverse backgrounds according to specified motion sequences. The key idea is to integrate the strengths of 3D human avatar-based modeling into diffusion-based frameworks, thereby enabling robust and realistic human-centric animation.

Specifically, since 3DGS human avatars possess view-consistent geometry through explicit 3D representations, their 2D renderings provide 3D-consistent structural information and a pose-aligned coarse appearance that remains stable across frames. Given a reference character image, the corresponding human avatar is first reconstructed, then animated and rendered according to the specified motion sequences. The rendered avatar frames are subsequently integrated with the associated scene video to form an avatar-background conditioning sequence, which closely approximates the ground-truth video except for local degradations around the human region, such as the loss of high-frequency textures or temporal dynamics in elements like hair and loose garments. The proposed integration process also naturally considers the spatial relationships between the character and the scene (\textit{e.g.}, occlusions). With this effective conditioning signal, the diffusion process is reformulated as a restoration task, \textit{i.e.}, 1) refining the degraded yet 3D-consistent appearance of human avatar renderings, and 2) synthesizing realistic non-rigid motion dynamics, such as the movement of hair and garments.

A dedicated data processing pipeline and a multi-conditional architecture are designed, incorporating a relighting augmentation strategy and an occlusion-aware avatar–scene integration mechanism to fully exploit the avatar-background conditioning signal while leveraging the rich prior knowledge embedded in the pre-trained base model. Consequently, the proposed approach combines the advantages of both diffusion-based and 3D avatar-based paradigms, resulting in superior visual generation quality. The codes and data will be released to facilitate further exploration.

\section{Related Work}

\begin{figure*}
  \includegraphics[width=.9\textwidth]{figures/sig_pipeline.jpeg}
  \setlength{\abovecaptionskip}{1mm}
  \caption{
    \textbf{Data Processing Pipeline.} 
    The video DiT model generates the output video by conditioning on three distinct signals: the reference image, the SMPL-X sequence, and the composite ``avatar--background'' video.
  }
  \label{fig:pipeline}
\end{figure*}

\noindent \textbf{Video diffusion models.}
Driven by recent advances in generative modeling~\cite{ho2020denoising,song2020denoising,ho2022video}, Video Diffusion Models (VDMs) have achieved remarkable progress in recent years~\cite{rombach2022high,kingma2013auto,van2017neural,esser2021taming,hacohen2024ltx,raffel2020exploring,radford2021learning,liu2023visual,li2024mini,ronneberger2015u,shi2026surprise,zhan2026composing}. Early approaches~\cite{he2022latent,xing2024dynamicrafter,chen2023videocrafter1,chen2024videocrafter2,blattmann2023stable} predominantly adopted UNet~\cite{ronneberger2015u}, whereas more recent efforts~\cite{hong2022cogvideo,yang2024cogvideox,li2024hunyuan,kong2024hunyuanvideo,wang2025wan} have transitioned toward Diffusion Transformers~\cite{peebles2023scalable}, which offer improved scalability and generation quality.

\noindent \textbf{Controllable video generation.}
Existing approaches employ a variety of control signals—such as trajectories~\cite{niu2024mofa,wang2024motionctrl,geng2024motion,lei2024animateanything,xu2024motion}, poses~\cite{hu2024animate,chang2025x,chang2023magicpose,niu2024mofa,zhu2024champ,zhang2024mimicmotion}, bounding boxes~\cite{huang2025fine,ma2024trailblazer,wang2024boximator,wu2024motionbooth,zhao2024motiondirector}, and camera parameters~\cite{bahmani2024ac3d,he2024cameractrl,yu2024viewcrafter}—to guide the generation process of VDMs. These techniques typically operate either by fine-tuning the base model~\cite{niu2024mofa,hu2024animate,xing2025motioncanvas,cao2024consistent,jiang2025vace} or through zero-shot conditioning~\cite{qi2023fatezero,qiu2024freetraj,khachatryan2023text2video,chen2024motion} without additional training.

\noindent \textbf{Character image animation.}
A variety of models~\cite{hu2024animate,zhang2024mimicmotion,wang2024unianimate,wang2024disco,xu2024magicanimate,zhu2024champ,lin2025omnihuman,hu2025animate,liu2025animateanywhere,men2024mimo,vace} have been developed to animate character images using diffusion-based frameworks according to specified motion sequences. These approaches leverage pre-trained foundation diffusion models to exploit extensive prior knowledge and directly synthesize 2D video sequences aligned with motion representations such as DWPose~\cite{yang2023effective}, DensePose~\cite{guler2018densepose}, or SMPL~\cite{loper2023smpl}. 
An alternative paradigm~\cite{qiu2025LHM,qiu2024anigs,wen2025lifegom,motionshop2} for character animation adopts explicit 3D modeling techniques, particularly 3D Gaussian Splatting~\cite{kerbl20233d}, to reconstruct a high-fidelity 3D avatar of the character before animating and rendering it into 2D images.

\begin{figure*}
\center
  \includegraphics[width=.9\textwidth]{figures/sig_architecture.jpeg}
  \caption{
    \textbf{Overall architecture.}
    To extract effective spatio-temporal representations, 3D convolutional layers and mask embeddings are incorporated to encode informative features from both the avatar-background video and the SMPL-X video.
  }
  \label{fig:architecture}
\end{figure*}

\begin{figure}
  \includegraphics[width=\linewidth]{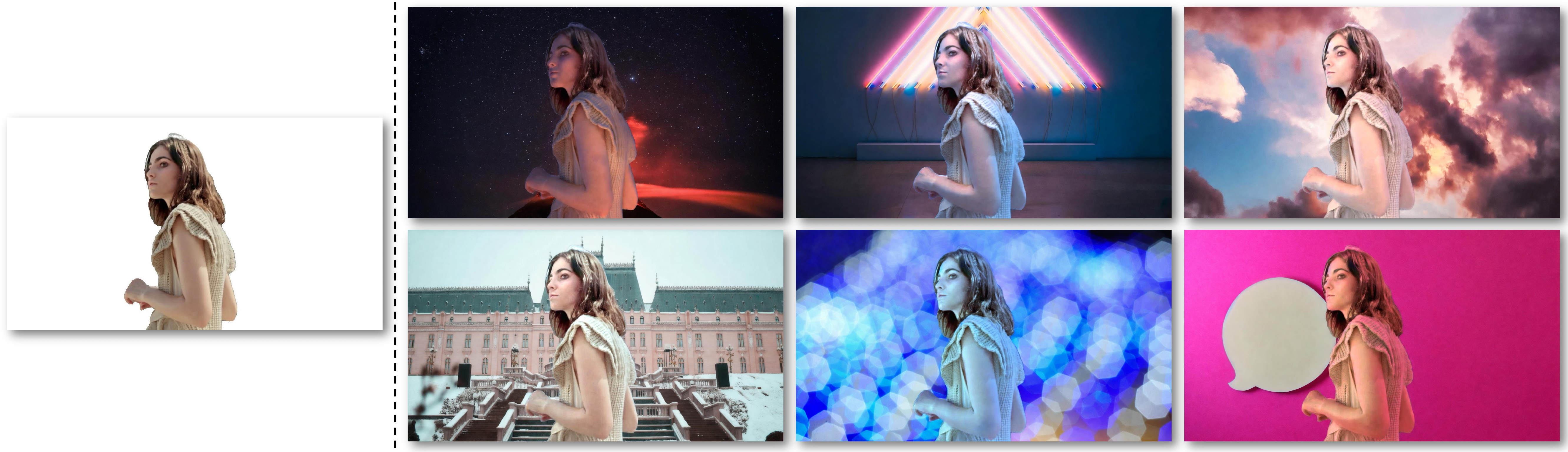}
  \caption{
    \noindent \textbf{Examples of relighting augmentation.} \textbf{Left:} Original cropped reference image. \textbf{Right:} Relighting augmentation results generated under different background conditions.
  }
  \label{fig:relighting}
\end{figure}

\section{Approach}
\label{sec:method}

\subsection{Training Paradigm}
\label{sec:pipeline}

\subsubsection{Overall pipeline}
\label{sec:overall_pipeline}

The data processing pipeline, illustrated in Fig.~\ref{fig:pipeline}, constructs high-quality training pairs from in-the-wild videos. Initially, SAM-2~\cite{ravi2024sam} generates accurate human segmentation masks to isolate the foreground, while ProPainter~\cite{zhou2023propainter} inpaints the resulting holes to produce temporally coherent background sequences. Simultaneously, monocular estimation recovers SMPL-X~\cite{loper2023smpl, SMPLX2019} parameters, which serve to animate 3D Gaussian Splatting (3DGS) avatars and provide the diffusion model with explicit structural motion cues via mesh visualizations.
To create the animatable avatar, LHM~\cite{qiu2025LHM} reconstructs a 3DGS representation from a randomly selected reference frame. This avatar is animated using the extracted SMPL-X sequence to produce pose-aligned renderings, and further
integrated with the backgrounds to form a composite ``avatar-background'' video. Ultimately, the diffusion model is trained using three inputs: the original reference image, the SMPL-X mesh sequence, and the avatar-background video.

\subsubsection{Relighting augmentation}
\label{sec:relighting}

Lighting inconsistencies pose a significant challenge in character animation. While standard training typically relies on self-reenactment pairs where the reference and background share identical illumination, real-world inference requires integrating arbitrary characters into environments with distinct lighting conditions. Consequently, strictly preserving the reference appearance can lead to unnatural mismatches in lighting and color tones.
To bridge this gap, a relighting augmentation strategy is introduced. Specifically, a reference character is first extracted from the original video. It is then composited onto a randomly chosen background using IC-Light~\cite{zhang2025scaling}, which automatically adjusts the illumination to align with the new environment.
As illustrated in Fig.~\ref{fig:relighting}, this process deliberately introduces lighting variations relative to the original sequence. This forces the network to learn effective lighting and color adaptation instead of mere texture copying, thereby achieving enhanced environmental harmonization during inference.

\subsubsection{Avatar–background integration}
\label{sec:blending}

To more accurately model the complex interactions between characters and their surrounding environments, two integration strategies are considered.

\noindent \textbf{Naive avatar–background alpha blending.}
The baseline case addresses scenarios without occlusions between the avatar and the scene. This configuration aligns with conventional human animation pipelines, and is particularly suited for applications that involve replacing or modifying the global background of a performing character (\textit{e.g.}, a dancing figure).
For non-occlusion cases, the animated human avatar sequence is composited with the inpainted background using alpha blending, guided by the Gaussian rasterizer’s opacity map to ensure smooth transitions. Formally, let the rendered avatar frame at time $t$ be denoted as $I^{\text{avatar}}_t \in \mathbb{R}^{H \times W \times 3}$, the inpainted background as $I^{\text{bg}}_t \in \mathbb{R}^{H \times W \times 3}$, and the opacity map generated by the Gaussian rasterizer as $\alpha_t \in [0,1]^{H \times W}$. The final composited frame is given by:
\begin{align}
I^{\text{final}}_t = \alpha_t \odot I^{\text{avatar}}_t + (1 - \alpha_t) \odot I^{\text{bg}}_t,
\end{align}
where $\odot$ denotes element-wise multiplication.

\noindent \textbf{Occlusion-aware avatar-scene integration.}
The advanced integration case handles complex scenarios involving significant occlusions between the character and surrounding scene objects by incorporating 3D geometric information from the scene video. For example, when a performer moves between a desk and a wall, the system must account for the occlusion caused by the desk to maintain realistic spatial relationships consistent with the scene geometry.
In such occlusion-sensitive settings, an occlusion mask $M^{\text{occ}}_t \in \{0,1\}^{H \times W}$ at time $t$, estimated from the scene geometry, is introduced to explicitly identify regions where the foreground avatar should be occluded by scene elements. This composition is defined as:
\begin{align}
I^{\text{final}}_t = M^{\text{occ}}_t \odot I^{\text{bg}}_t + (1 - M^{\text{occ}}_t) \odot \big( \alpha_t \odot I^{\text{avatar}}_t + (1 - \alpha_t) \odot I^{\text{bg}}_t \big),
\end{align}
where $\odot$ denotes element-wise multiplication.
This formulation preserves realistic occlusion relationships between the avatar and the scene: regions corresponding to $M^{\text{occ}}_t = 1$ are replaced with the original background $I^{\text{bg}}_t$, ensuring that closer objects fully occlude the avatar. In the remaining regions where $M^{\text{occ}}_t = 0$, the avatar and background are blended using the Gaussian rasterizer's transparency map $\alpha_t$, achieving visually coherent integration.

To estimate the occlusion mask from the source video, SAM-2 is utilized to obtain semantic segmentation masks for each frame, including the human mask. Let $\{M^{(k)}_t\}_{k=1}^K$ represent the set of semantic masks extracted from frame $t$, where $M^{\text{human}}_t$ corresponds to the human region. A video depth estimator~\cite{yang2024depth} is then applied to predict a per-frame depth map $D_t \in \mathbb{R}^{H \times W}$. For each semantic region $M^{(k)}_t$, its mean depth is computed as:
\begin{equation}
d^{(k)}_t = \frac{\sum_{i,j} D_t(i,j) \cdot M^{(k)}_t(i,j)}{\sum_{i,j} M^{(k)}_t(i,j)},
\label{eq:avg-depth}
\end{equation}
where $(i,j)$ denote pixel coordinates. The occlusion mask $M^{\text{occ}}_t$ is subsequently defined as the union of all semantic regions whose mean depth is smaller than that of the human region:
\begin{equation}
M^{\text{occ}}_t = \bigcup_{k \neq \text{human}} \mathbbm{1} \left[ d^{(k)}_t < d^{\text{human}}_t \right] \cdot M^{(k)}_t,
\label{eq:occ-mask}
\end{equation}
where $\mathbbm{1}[\cdot]$ denotes the indicator function. This formulation ensures that any object located closer to the camera than the human is identified as an occluder and incorporated into the occlusion mask.

\vspace{-2mm}
\subsection{Diffusion Model Architecture}
\label{sec:architecture}

\noindent \textbf{Overall architecture.}
Fig.~\ref{fig:architecture} presents the architecture of the proposed diffusion model. Building upon Wan2.1-I2V~\cite{wang2025wan}, the model is designed to predict the combination of ``reference image + target video'' rather than the ``target video'' alone. This design choice provides two main advantages: (1) it preserves structural consistency with standard I2V diffusion frameworks, ensuring that the first generated frame corresponds to the input reference image---a crucial property when the reference image does not align with the initial frame of the target video; and (2) it enables the proposed mask embedding strategy to effectively exploit both the auxiliary conditioning inputs and the generative priors of the pre-trained Wan2.1-I2V. During training, we integrate Low-Rank Adaptation (LoRA)~\cite{hu2022lora} layers into each Diffusion Transformer (DiT) block while keeping the original network parameters frozen to preserve the pre-trained model's stability and generalization capabilities.

As illustrated in Fig.~\ref{fig:architecture}, instead of encoding only the single reference image into the latent space, the proposed model encodes the frame-wise concatenation of the reference image and the target video. A dedicated mask embedding strategy is then applied to the encoded latent, ensuring strict compatibility with the original Wan2.1 mask design. This architectural alignment facilitates the effective utilization of pre-trained generative priors, while the modification enables the adaptive integration of auxiliary information from the SMPL-X and avatar-background conditioning sequences.
Inspired by Unianimate-DiT~\cite{wang2025unianimate}, a series of 3D convolutional layers are employed to extract spatio-temporal features from both the SMPL-X mesh videos and the avatar-background videos. The inclusion of SMPL-X mesh videos serves two purposes: (1) providing regional guidance for the mask embedding strategy, and (2) supplying pure structural information explicitly decoupled from the avatar-background condition.

\begin{figure}
  \includegraphics[width=\linewidth]{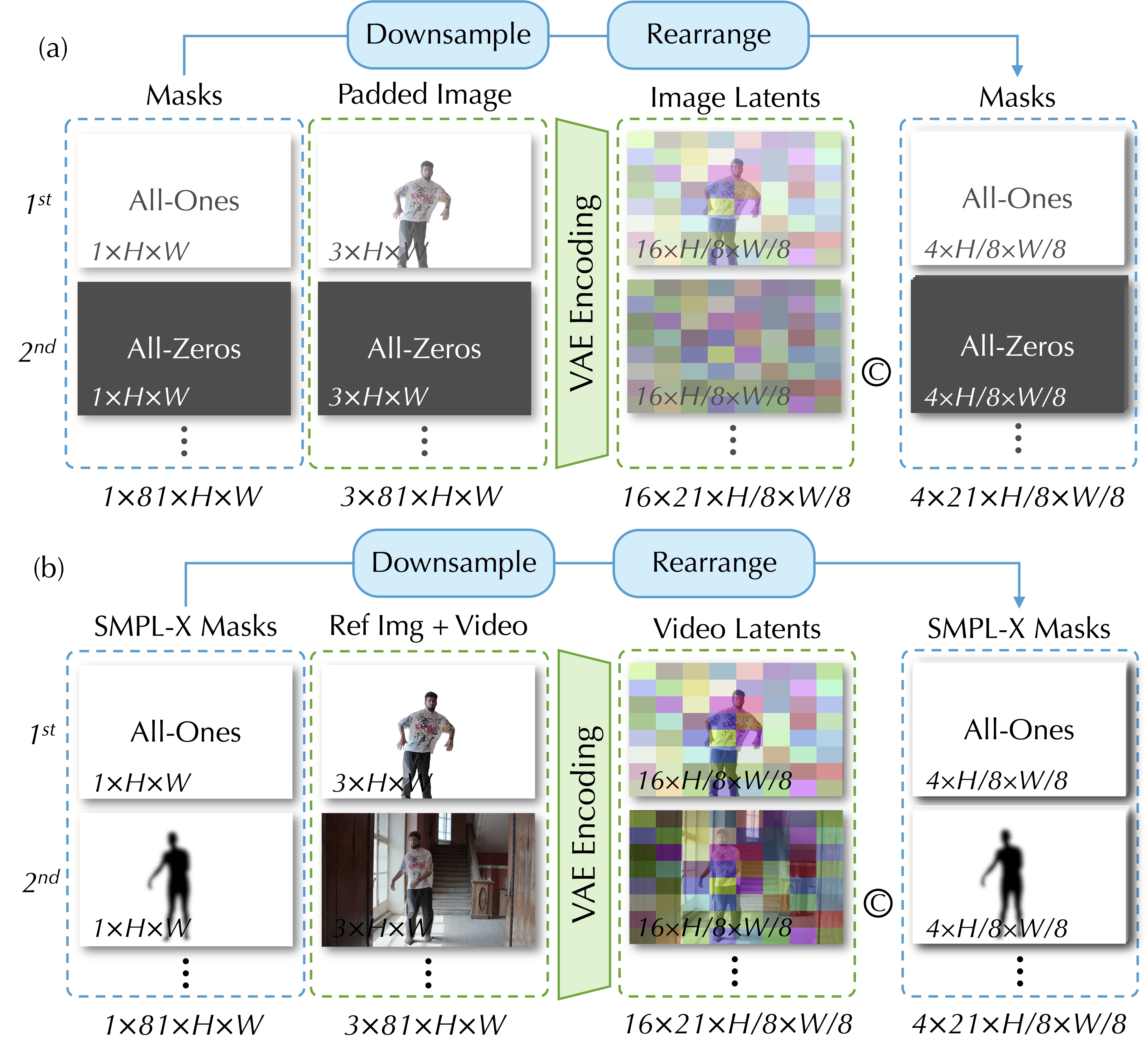}
  \setlength{\abovecaptionskip}{-2mm}
  \caption{
    \textbf{Illustration of the mask strategy} for (a) Wan2.1-I2V~\cite{wang2025wan} and (b) the proposed method. 
  }
  \vspace{-4mm}
  \label{fig:mask_strategy}
\end{figure}

\noindent \textbf{Mask embedding strategy.}
In Wan2.1-I2V, the input image $\mathbf{I}_0 \in \mathbb{R}^{3 \times H \times W}$ is first zero-padded to match the temporal dimension of the target video $\mathbf{V} \in \mathbb{R}^{3 \times 81 \times H \times W}$ and then encoded into the latent space through the Causal VAE, yielding a latent representation $\mathbf{L} \in \mathbb{R}^{16 \times 21 \times H/8 \times W/8}$. As illustrated in Fig.~\ref{fig:mask_strategy}~(a), the framework introduces a binary mask $\mathbf{M} \in \{0, 1\}^{1 \times 81 \times H \times W}$, where a mask value of $1$ corresponds to preserved content (i.e., the input image in Wan2.1-I2V) and $0$ indicates frames to be generated.
This mask is processed in two stages: first, it is temporally downsampled to align with the temporal resolution of the latent representation $\mathbf{L}$; second, the all-one mask of the first frame is repeated four times to form the final mask representation $\mathbf{M}_{L} \in \{0, 1\}^{4 \times 21 \times H/8 \times W/8}$.

Following the principle of using the diffusion model as a ``refinement'' module guided by the avatar–background condition, a compatible mask embedding strategy is designed to fully incorporate information from the ``avatar–background'' video while retaining the logic of Wan2.1-I2V. As shown in Fig.~\ref{fig:mask_strategy}~(b), instead of encoding a zero-padded reference image $\mathbf{I}$, the model encodes the 
frame-wise concatenation of the reference image $\mathbf{I} \in \mathbb{R}^{3 \times 1 \times H \times W}$ and the ``avatar–background'' video $\mathbf{V}_r \in \mathbb{R}^{3 \times 80 \times H \times W}$ as:
\begin{align}
\hat{\mathbf{L}} = \mathcal{E}(\operatorname{concat}[\mathbf{I}, \mathbf{V}_r]) \in \mathbb{R}^{16 \times 21 \times H/8 \times W/8}.
\end{align}
In Wan2.1-I2V, a mask value of $1$ denotes preserved frames (input image), whereas $0$ indicates frames to be generated. This work extends the concept into a spatially variant formulation by introducing 0–1 masks derived from SMPL-X meshes to identify human body regions. Because SMPL-X mask boundaries may not perfectly coincide with the true body contours, Gaussian blur is applied to produce soft transition edges. The diffusion model thus focuses on refining the 3DGS avatar-rendered human body regions while maintaining background areas that already align with the ground-truth video. Consequently, mask values are assigned as $0$ for body regions (requiring refinement) and $1$ for background regions (to be preserved). Occluded regions are also set to $1$ according to the occlusion map described in Sec.~\ref{sec:blending}.
This enhanced mask strategy enables the model to leverage the pretrained priors of the base model while concentrating refinement on the human body areas.

\begin{table}[t]
\centering
\setlength{\tabcolsep}{2mm}
\setlength{\abovecaptionskip}{1mm}
\caption{\textbf{User study} for cross ID-background evaluation. Results are annotated with \textbf{the best} and \underline{the second best}.}
\label{tab:user_study}
\resizebox{\linewidth}{!}{%
\begin{tabular}{lcccccc}
\toprule
Model & \small{ID Preserv.} & \small{Body Struc.} & \small{Motion Dyna.} & \small{Overall Quali.} \\
\midrule
HumanVid & 2.32 & 2.47 & 2.11 & 2.44  \\
MIMO & 2.51 & 2.35 & 1.94 & 2.93 \\
VACE & \underline{3.35} & \underline{3.28} & \underline{3.44} & \underline{3.31} \\
Ours & \textbf{4.02} & \textbf{4.25} & \textbf{3.91} & \textbf{4.33}  \\
\bottomrule
\end{tabular}%
}
\end{table}

\begin{table}[t]
\centering
\setlength{\tabcolsep}{2.2mm}
\setlength{\abovecaptionskip}{1mm}
\caption{\textbf{Quantitative comparison} on the human dancing test set. Results are annotated with \textbf{the best} and \underline{the second best}.}
\label{tab:dancing}
\resizebox{\linewidth}{!}{%
\begin{tabular}{lccccc}
\toprule
Methods & SSIM $\uparrow$ & PSNR $\uparrow$ & LPIPS $\downarrow$ & FID $\downarrow$ & FVD $\downarrow$  \\ 
\midrule
Champ & 0.6678 & 15.595 & 0.3182 & 37.402 & 1365.24 \\
StableAnimator & 0.8216 & 20.830 & 0.1856 & 26.435 & 590.87 \\
Unianimate-DiT & {\underline{0.8726}} & {\underline{23.153}} & {\underline{0.1202}} & {\underline{17.201}}  & {\underline{305.69}} \\
Ours & \textbf{0.8870} & \textbf{23.992} & \textbf{0.1035}  & \textbf{12.957} & \textbf{253.68} \\
\bottomrule
\end{tabular}%
}
\end{table}

\begin{table}[t]
\centering
\setlength{\tabcolsep}{3mm}
\setlength{\abovecaptionskip}{1mm}
\caption{\textbf{Quantitative comparison} on the HumanVid test set. Results are annotated with \textbf{the best} and \underline{the second best}.}
\label{tab:humanvid}
\resizebox{\linewidth}{!}{%
\begin{tabular}{lccccc}
\toprule
Methods & SSIM $\uparrow$ & PSNR $\uparrow$ & LPIPS $\downarrow$ & FID $\downarrow$ & FVD $\downarrow$  \\ 
\midrule
HumanVid & 0.7316 & 21.559 & 0.2035 & 61.584 & {{676.98}} \\
MIMO & {\underline{0.8737}} & {\underline{23.014}} & {\underline{0.1158}} & {{37.074}} & 768.79 \\ 
VACE & {{0.7934}} & {{22.786}} & {{0.1355}} & {\underline{23.989}} & \underline{288.25} \\ 
Ours & \textbf{0.9091} & \textbf{27.681} & \textbf{0.0721} & \textbf{9.276} & \textbf{259.06} \\
\bottomrule
\end{tabular}%
}
\end{table}

\begin{figure*}[t]
\centering
    \setlength{\abovecaptionskip}{0mm}
    \includegraphics[width=\textwidth]{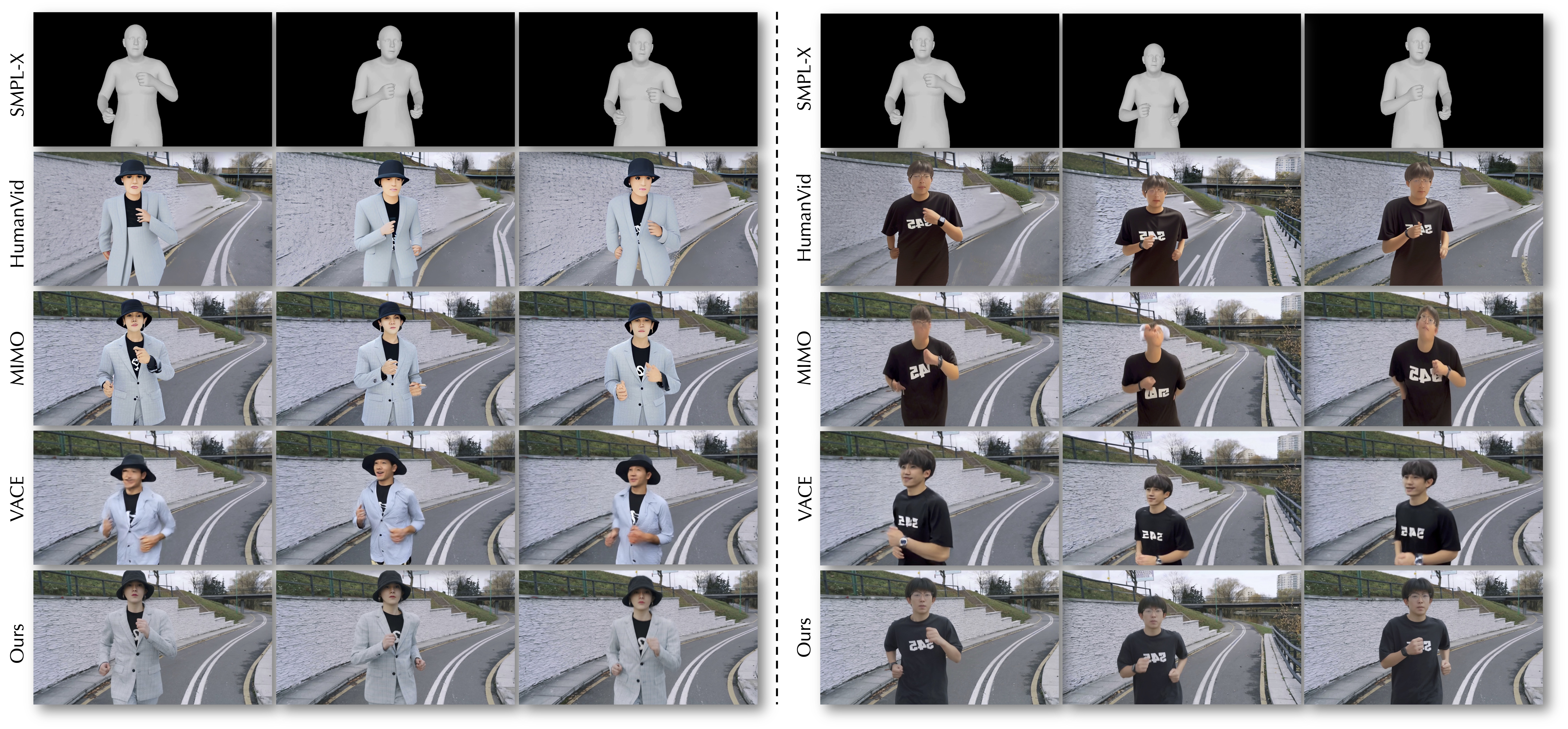}
    \captionof{figure}{
    \textbf{Qualitative comparisons for cross character–background evaluation.} The pose of each frame is visualized as a motion reference. Zoom in for the best view. 
    }
    \label{fig:cross}
\end{figure*}

\begin{figure*}[t]
\centering
    \setlength{\abovecaptionskip}{-1mm}
    \includegraphics[width=\textwidth]{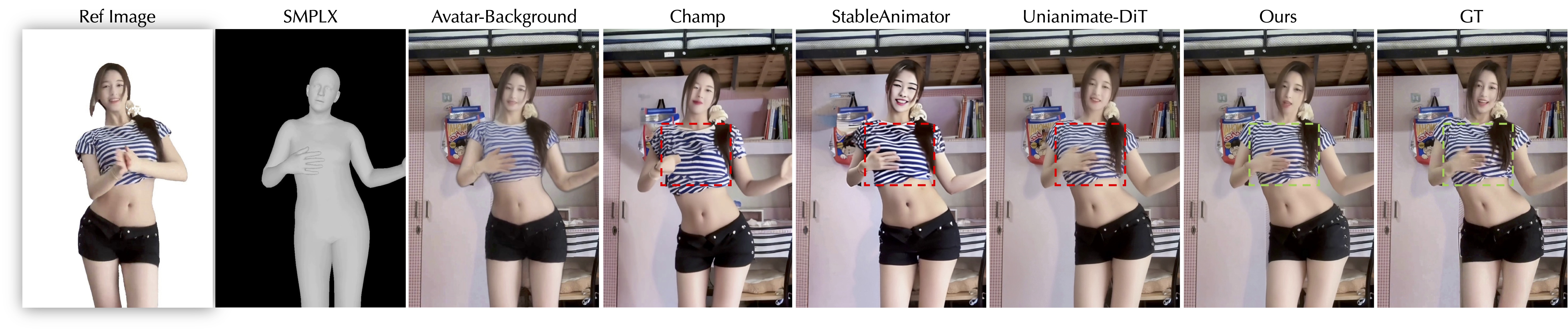}
    \captionof{figure}{
    \textbf{Self-reenactment results on the Dancing dataset.} 
    }
    \label{fig:dancing}
\end{figure*}

\subsection{Implementation Details}
\label{sec:implement}

The model is first trained on eight NVIDIA H200 GPUs using a self-collected human dancing dataset containing approximately 3,000 videos. It is then fine-tuned on a curated subset of around 5,000 video clips selected from HumanVid~\cite{wang2024humanvid}. The self-collected dataset primarily consists of videos with static backgrounds, while the filtered HumanVid dataset includes dynamic background scenes. This hybrid training strategy facilitates the learning of both human motion and background dynamics.
Wan2.1-I2V-14B~\cite{wang2025wan} is adopted as the base model for weight initialization. The model is optimized using AdamW~\cite{loshchilov2017decoupled} for 15,000 iterations with a learning rate of $1 \times 10^{-4}$. Although training is conducted at a spatial resolution of 480p, the model exhibits strong generalization capabilities when evaluated on 720p inputs.

\section{Experiments}
\label{sec:experiment}

\subsection{Cross Character-Background Evaluation}
\label{sec:user_study}

The proposed method is evaluated on cross character–background samples, where 10 in-the-wild character images are animated using 20 different dynamic background videos following given motion sequences. Since ground-truth videos are unavailable, a user study and extensive qualitative evaluations are conducted. Volunteers are asked to rate the generated videos on a scale of 1 to 5 (higher is better) based on four criteria: (1) identity preservation, (2) body structure, (3) secondary motion dynamics, and (4) overall quality.

\begin{table}[t]
\centering
\setlength{\abovecaptionskip}{2mm}
\caption{\textbf{Quantitative ablation} on network architecture. Results are annotated with \textbf{the best} and \underline{the second best}.}
\label{tab:ablation}
\resizebox{\linewidth}{!}{%
\begin{tabular}{llccccc}
\toprule
Dataset & Model & SSIM $\uparrow$ & PSNR $\uparrow$ & LPIPS $\downarrow$ & FID $\downarrow$ & FVD $\downarrow$ \\
\midrule
\multirow{3}{*}{Dancing} & w/o Avatar & 0.8757 & 23.267 & 0.1154 & 14.194 & 262.23 \\
 & w/o Mask & \underline{0.8853} & \underline{23.776} & \underline{0.1073} & \underline{13.723} & \underline{257.11} \\
& Ours & \textbf{0.8870} & \textbf{23.992} & \textbf{0.1035}  & \textbf{12.957} & \textbf{253.68} \\
\midrule
\multirow{3}{*}{HumanVid} & w/o Avatar & 0.9018 & 27.246 & 0.0772 & 10.651 & 274.69 \\
 & w/o Mask & \underline{0.9052} & \textbf{27.692} & \textbf{0.0706} & \underline{10.256} & \underline{261.07} \\
& Ours & \textbf{0.9091} & \underline{27.681} & \underline{0.0721} & \textbf{9.276} & \textbf{259.06} \\
\bottomrule
\end{tabular}%
}
\end{table}

\begin{figure*}[t]
\centering
    \setlength{\abovecaptionskip}{1mm}
    \includegraphics[width=\textwidth]{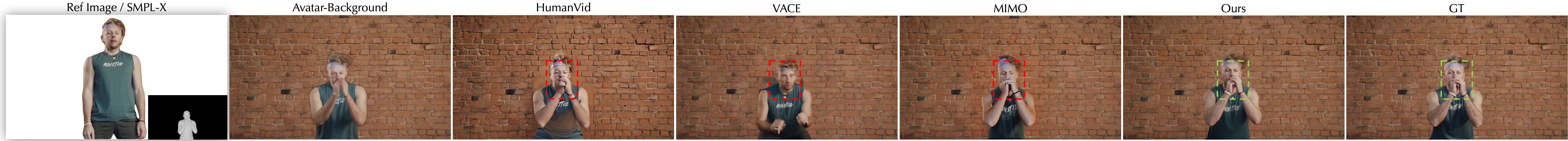}
    \captionof{figure}{
    \textbf{Self-reenactment results on the HumanVid dataset.} 
    }
    \label{fig:humanvid}
\end{figure*}

\begin{figure*}[t]
\centering
    \setlength{\abovecaptionskip}{0mm}
    \includegraphics[width=\linewidth]{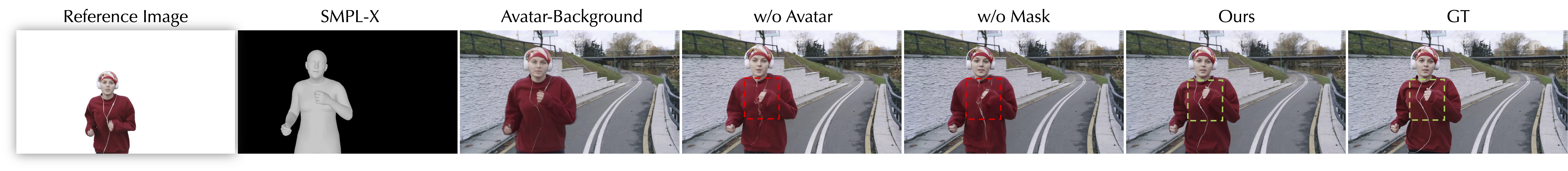}
    \captionof{figure}{
    \textbf{Qualitative ablation results for our model.} 
    }
    \label{fig:ablation}
\end{figure*}

Comparisons are made with HumanVid~\cite{wang2024humanvid} [NeurIPS'24], MIMO \cite{men2024mimo} [CVPR'25], and VACE~\cite{vace} [ICCV'25], considering both background dynamics and character animation. For HumanVid~\cite{wang2024humanvid}, the parsed character image is manually composited with the background to form the reference input. For MIMO~\cite{men2024mimo} and VACE~\cite{vace}, the official data processing pipelines are followed to produce the animation results. The user study outcomes are reported in Table~\ref{tab:user_study}, and qualitative comparisons are shown in Fig.~\ref{fig:cross}.
Compared with HumanVid, MIMO, and VACE, the proposed method achieves superior ID preservation, improved structural stability, and fewer visual artifacts. The accompanying video further illustrates that the method produces more realistic motion dynamics for both the human body and non-rigid elements, such as loose garments. The user study results confirm that participants consistently favor the proposed approach across all evaluation aspects.

\subsection{Self-Reenactment Evaluation}
\label{sec:self_reenactment}

\begin{figure}[t]
\centering
    \setlength{\abovecaptionskip}{2mm}
    \includegraphics[width=\linewidth]{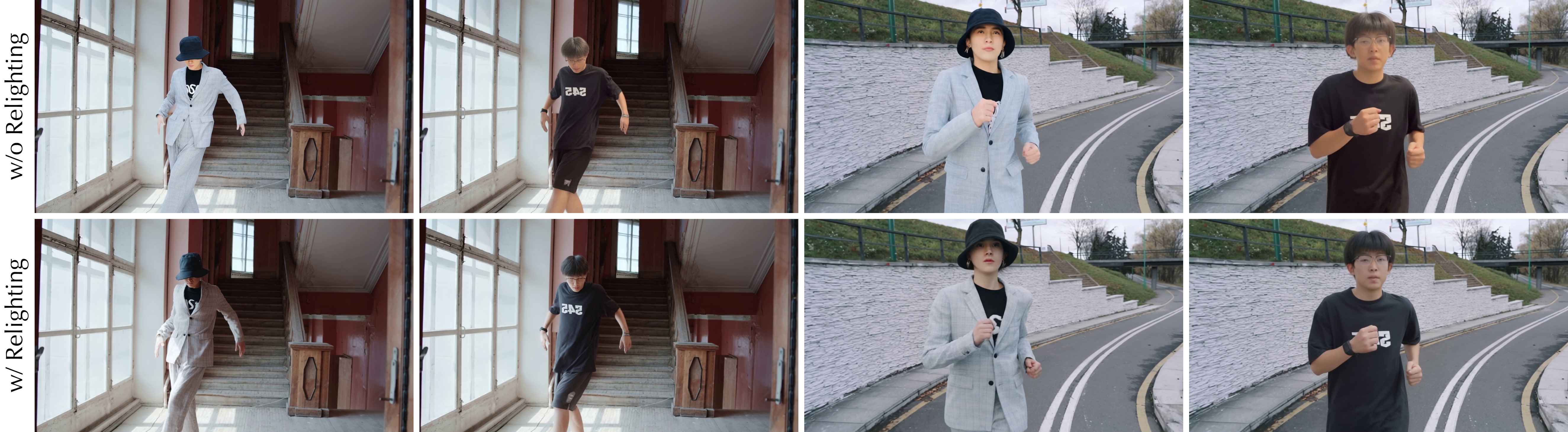}
    \captionof{figure}{
    \textbf{Qualitative ablation results for relighting augmentation.} 
    }
    \label{fig:relighting_ablation}
\end{figure}

\begin{table}[t]
\centering
\setlength{\tabcolsep}{2mm}
\setlength{\abovecaptionskip}{2mm}
\caption{\textbf{User study} for relighting augmentation.}
\label{tab:user_study_relighting}
\resizebox{.75\linewidth}{!}{%
\begin{tabular}{lcc}
\toprule
Model & Lighting Harmony & Overall Quality \\ \midrule
w/o Relighting & 3.24 & 3.91 \\
w/ Relighting & \textbf{4.42} & \textbf{4.28} \\ \bottomrule
\end{tabular}
}
\end{table}

\noindent \textbf{Static backgrounds.}
The proposed method is first compared with state-of-the-art character animation approaches on self-reenactment tasks. For a fair quantitative evaluation, 60 test videos \textit{with static backgrounds and no character-scene occlusions} are selected from the human dancing dataset for metric computation. Since most comparison methods do not model background motion and assume that the video background matches the reference image, including videos with dynamic backgrounds and occlusions would lead to an unfair comparison.
Specifically, comparisons are conducted with the following methods: (1) Champ~\cite{zhu2024champ} [ECCV'24], which conditions generation on 3D information such as SMPL parameters, normal maps, and depth; (2) StableAnimator~\cite{tu2024stableanimator} [CVPR'25], which performs animation using DWPose~\cite{yang2023effective} sequences; and (3) Unianimate-DiT~\cite{wang2025unianimate} [arXiv'25]. Unianimate-DiT is included as it is the SOTA open-source character animation model that shares the same base model, Wan2.1-I2V-14B. The following metrics are computed between the generated outputs and the original video clips: SSIM, PSNR, LPIPS~\cite{zhang2018unreasonable}, FID~\cite{heusel2017gans}, and FVD~\cite{unterthiner2018towards}. Quantitative results are reported in Table~\ref{tab:dancing}, where the proposed method achieves the best performance across all metrics. Qualitative comparisons are shown in Fig.~\ref{fig:dancing}. For Champ, StableAnimator, and Unianimate-DiT, the reference images are used with their original backgrounds, following their respective standard practices. The proposed method achieves better structural fidelity with fewer visual artifacts. 

\begin{figure}[t]
\centering
    \setlength{\abovecaptionskip}{2mm}
    \includegraphics[width=\linewidth]{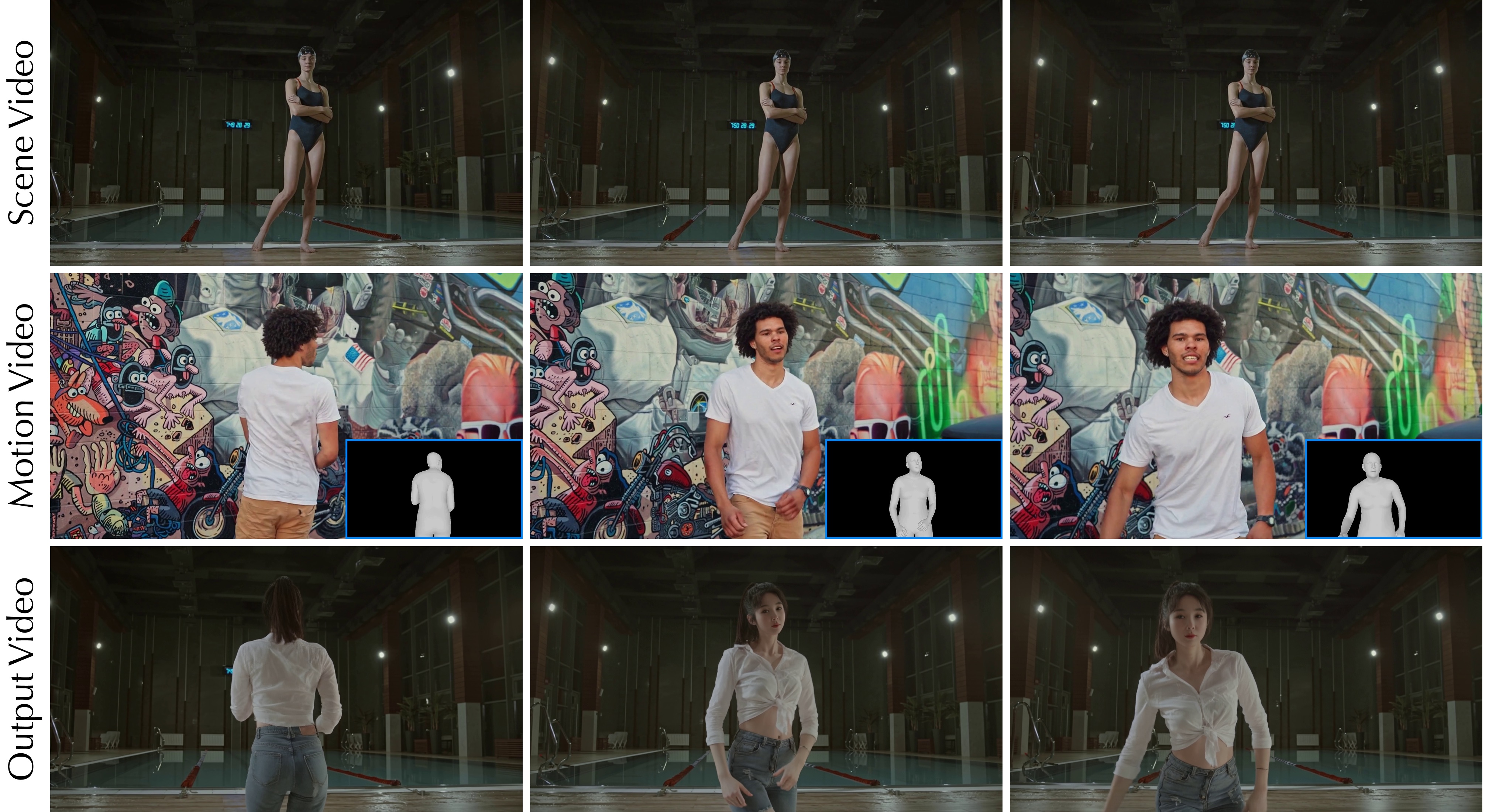}
    \captionof{figure}{
    \textbf{Qualitative novel motion animation results.} 
    }
    \label{fig:novel_motion}
\end{figure}

\begin{figure*}[t]
\centering
    \setlength{\abovecaptionskip}{2mm}
    \includegraphics[width=\linewidth]{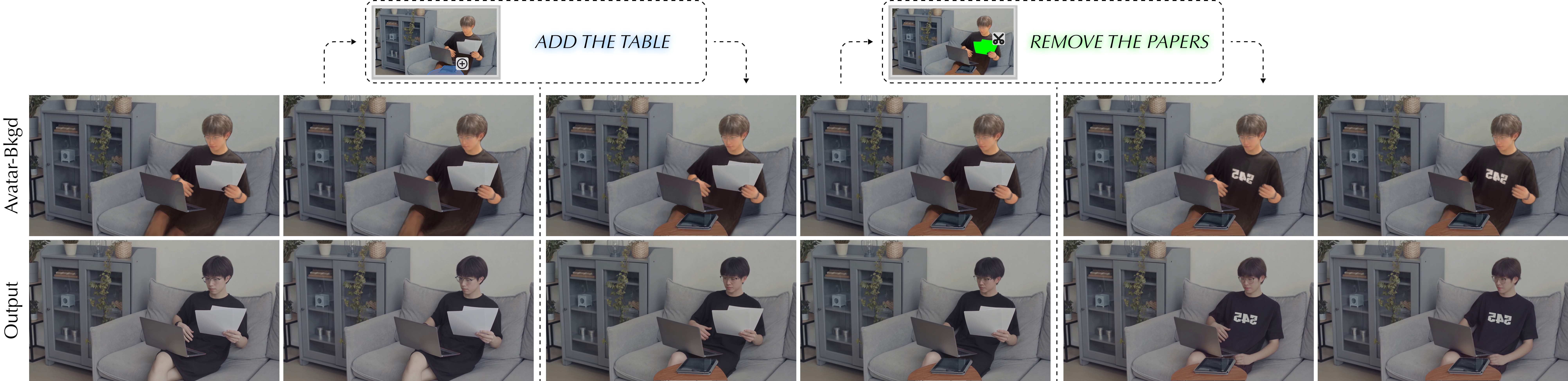}
    \captionof{figure}{
    \textbf{Qualitative results for occlusion-aware editing.} 
    }
    \label{fig:occ_edit}
\end{figure*}

\begin{figure*}[t]
\centering
    \setlength{\abovecaptionskip}{2mm}
    \includegraphics[width=\linewidth]{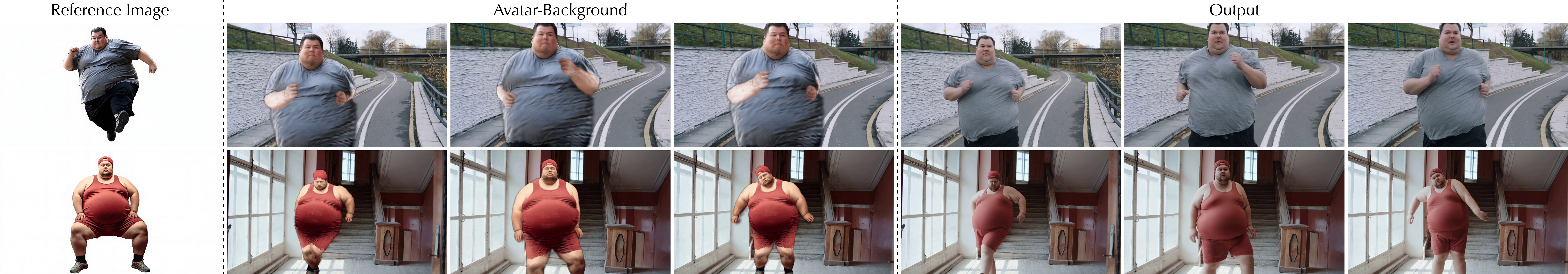}
    \captionof{figure}{
    \textbf{Qualitative animation results for various body shapes.} 
    }
    \label{fig:novel_shape}
\end{figure*}

\noindent \textbf{Open-domain dynamic backgrounds.}
For self-reenactment evaluation under dynamic background conditions, 100 test videos are selected from the filtered HumanVid dataset. Comparisons are conducted with HumanVid~\cite{wang2024humanvid} [NeurIPS'24], MIMO~\cite{men2024mimo} [CVPR'25], and VACE~\cite{vace} [ICCV'25]. Both quantitative and qualitative results are presented in Table~\ref{tab:humanvid} and Fig.~\ref{fig:humanvid}, respectively.
It is noteworthy that MIMO does not achieve a lower FVD score than HumanVid, even though it leverages dynamic background videos as inputs. This outcome can be attributed to the characteristics of the FVD metric, which emphasizes the naturalness of temporal video dynamics---such as secondary motion and the realism of human movement---rather than pixel-level reconstruction accuracy. While MIMO improves image-level metrics by incorporating dynamic background inputs, it struggles to synthesize temporally coherent and realistic motion, which is a key factor in human animation quality.
In contrast, the proposed method achieves superior performance by jointly modeling character motion and background dynamics through the avatar--background conditioning mechanism.

\subsection{Component Evaluation}
\label{sec:ablation}

\noindent \textbf{Network architecture.}
To assess the contribution of individual components, comprehensive ablation studies are conducted on the diffusion model architecture. Specifically, the full model is compared against two variants: (1) \textit{w/o Avatar}, which removes the ``avatar--background'' encoding and relies solely on the background video and SMPL-X mesh sequences, and (2) \textit{w/o Mask}, which excludes the mask embedding strategy described in Section~\ref{sec:architecture}. Quantitative and qualitative results are reported in Table~\ref{tab:ablation} and Fig.~\ref{fig:ablation}, respectively. The results indicate that the complete model---integrating both 3D-consistent human avatar rendering and the proposed mask strategy---achieves the best overall performance. 

\noindent \textbf{Relighting augmentation.}
As illustrated in Fig.~\ref{fig:relighting_ablation}, without relighting augmentation, the generated characters often exhibit inconsistent illumination, where their skin tones and clothing reflections remain static regardless of the environment. This is particularly noticeable in scenes with strong directional light or specific color temperatures. In contrast, our full model effectively adapts the character's appearance to the background's global illumination. 
We further performed a user study under a similar protocol to that described in Sec.~\ref{sec:user_study}, focusing on two key aspects: Lighting Harmony and Overall Quality. As shown in Table~\ref{tab:user_study_relighting}, the inclusion of relighting augmentation leads to a substantial gain in both metrics, confirming the effectiveness of our relighting strategy.

\subsection{Further Evaluations}
\label{sec:further_evaluation}

\noindent \textbf{Novel motion.}
While the primary experiments focus on reenactment and cross character-background scenarios, the motion sequences in these settings are typically coupled with their corresponding source videos. Fig.~\ref{fig:novel_motion} evaluates the generality of the model by applying novel motion sequences to the original scene. The model successfully transfers motion patterns from a different video while preserving visual coherence within the original scene.

\noindent \textbf{Different body shapes.}
In some cases, the target character has a significantly different body shape from the neutral body or the subject in the original video. Animation results under such extreme body shape are shown in Fig.~\ref{fig:novel_shape}. The model maintains both motion fidelity and visual realism under these extreme conditions. 

\noindent \textbf{Limitations.} Currently, our model does not incorporate explicit dynamic 4D modeling for scene elements. This limitation prevents the model from being applied to tasks such as 4D-aware novel view synthesis. A promising direction would involve the joint reconstruction and animation of both human subjects and background elements using explicit 4D Gaussian Splatting representations.

\section{Conclusion}
\label{sec:conclusion}

This paper presents a human-centric animation model capable of seamlessly inserting and animating subjects into arbitrary open-domain dynamic backgrounds while preserving realistic motion dynamics according to specified motion sequences. Our approach combines an avatar--background conditioning paradigm, which provides robust and occlusion-aware guidance, with a diffusion-based architecture for high-fidelity video generation. Extensive experimental evaluations demonstrate the superior performance of the proposed method compared to existing alternatives. Future work involves the joint animation of both human subjects and background elements via 4D Gaussian Splatting representations.

\section*{Acknowledgments}
This work was supported in part by JSPS KAKENHI Grant Number 24KK0209, the Forest Digital Twin Project under the Partnership Agreement for Social Value Creation between UTokyo and SMBC Group, the Hokkaido Sarabetsu Village ``Endowed Chair for Field Phenomics'' projects in Japan, and the Advanced AI Talent Development to Lead the Next-Generation AI for Intelligent Society (BOOST NAIS) of The University of Tokyo.

\bibliographystyle{ACM-Reference-Format}
\bibliography{sample-base}

\end{document}